\documentclass[twoside]{article}
\usepackage[accepted]{arxiv}

\newcommand{\inner}[2]{\langle #1,#2\rangle}

\newcommand{\boldone}{{\boldsymbol{1}}}

\newcommand{\boldA}{{\boldsymbol{A}}}

\newcommand{\boldH}{{\boldsymbol{H}}}
\newcommand{\boldI}{{\boldsymbol{I}}}

\newcommand{\boldK}{{\boldsymbol{K}}}
\newcommand{\boldL}{{\boldsymbol{L}}}

\newcommand{\boldW}{{\boldsymbol{W}}}

\newcommand{\boldf}{{\boldsymbol{f}}}
\newcommand{\boldg}{{\boldsymbol{g}}}
\newcommand{\boldh}{{\boldsymbol{h}}}

\newcommand{\boldk}{{\boldsymbol{k}}}

\newcommand{\boldx}{{\boldsymbol{x}}}
\newcommand{\boldy}{{\boldsymbol{y}}}

\newcommand{\boldalpha}{{\boldsymbol{\alpha}}}

\newcommand{\boldGamma}{{\boldsymbol{\Gamma}}}

\newcommand{\boldPi}{{\boldsymbol{\Pi}}}
\newcommand{\boldSigma}{{\boldsymbol{\Sigma}}}

\usepackage{amsmath,epsfig,bm, amssymb,subfigure,natbib}

\begin{document}

%

%

\twocolumn[

\aistatstitle{Cross-Domain Object Matching with Model Selection}



\aistatsauthor{Makoto Yamada \And Masashi Sugiyama }

\aistatsaddress{ Tokyo Institute of Technology\\ \tt{yamada@sg.cs.titech.ac.jp} \And Tokyo Institute of Technology \\ \tt{sugi@cs.titech.ac.jp}} ]

\begin{abstract}
 The goal of \emph{cross-domain object matching} (CDOM) is to find correspondence between two sets of objects in different domains in an unsupervised way. Photo album summarization is a typical application of CDOM, where photos are automatically aligned into a designed frame expressed in the Cartesian coordinate system. CDOM is usually formulated as finding a mapping from objects in one domain (photos) to objects in the other domain (frame) so that the pairwise dependency is maximized. A state-of-the-art CDOM method employs a kernel-based dependency measure, but it has a drawback that the kernel parameter 
needs to be determined manually. In this paper, we propose alternative CDOM methods that can naturally address the model selection problem.  Through experiments on image matching, unpaired voice conversion, and photo album summarization tasks, the effectiveness of the proposed methods is demonstrated. 
\end{abstract}

\section{Introduction}
The objective of \emph{cross-domain object matching} (CDOM) is to match two sets of objects in different domains. For instance, in photo album summarization, photos are automatically assigned into a designed frame expressed in the Cartesian coordinate system. A typical approach of CDOM is to find a mapping from objects in one domain (photos) to objects in the other domain (frame) so that the pairwise dependency is maximized.
In this scenario, accurately evaluating the dependence between objects is a key challenge.

\emph{Kernelized sorting} (KS) \citep{COLT:Jebara:2004} tries to find a mapping
between two domains that maximizes the \emph{mutual information} (MI)
\citep{book:Cover+Thomas:2006} under the Gaussian assumption.
However, since the Gaussian assumption may not be fulfilled in practice,
this method (which we refer to as KS-MI) tends to perform poorly.

To overcome the limitation of KS-MI,
\citet{PAMI:Quadrianto+etal:2010} proposed using
the kernel-based dependence measure
called the \emph{Hilbert-Schmidt independence criterion} (HSIC) \citep{ALT:Gretton+etal:2005}
for KS.
Since HSIC is distribution-free,
KS with HSIC (which we refer to as KS-HSIC) is more flexible than KS-MI.
However, HSIC includes a tuning parameter (more specifically, the Gaussian kernel width),
and its choice is crucial to obtain better performance
\citep[see also][]{AAAI:Hal+etal:2010}.
Although using the median distance between sample points as the Gaussian kernel width
is a common heuristic in kernel-based dependence measures
\citep[see e.g.,][]{TAS:Fukumizu+etal:2009},
this does not always perform well in practice.

In this paper, we propose two alternative CDOM methods that can naturally address
the model selection problem.
The first method employs another kernel-based dependence measure based on
the \emph{normalized cross-covariance operator} (NOCCO) \citep{NIPS:Fukumizu+etal:2008},
which we refer to as KS-NOCCO. 
The NOCCO-based dependence measure was shown to be
asymptotically independent of the choice of kernels.
Thus, KS-NOCCO is expected to be less sensitive to the kernel parameter choice,
which is an advantage over HSIC.

The second method uses \emph{least-squares mutual information} (LSMI) \citep{BMCBio:Suzuki+etal:2009a}
as the dependence measure, which is a consistent estimator of 
the \emph{squared-loss mutual information} (SMI) achieving the optimal convergence rate.
We call this method \emph{least-squares object matching} (LSOM).
An advantage of LSOM is that 
cross-validation (CV) with respect to the LSMI criterion is possible.
Thus, all the tuning parameters
such as the Gaussian kernel width and the regularization parameter
can be objectively determined by CV.

Through experiments on image matching, unpaired voice conversion, and photo album summarization tasks, 
LSOM is shown to be the most promising approach to CDOM.
\section{Problem Formulation} \label{sec:prob_formulate}
In this section, we formulate the problem of \emph{cross-domain object matching} (CDOM).

The goal of CDOM is, given two sets of samples of the same size,
$\{\boldx_i\}_{i=1}^n$ and $\{\boldy_i\}_{i=1}^n$,
to find a mapping that well ``matches'' them.

Let $\pi$ be a permutation function over $\{1,\ldots,n\}$,
and let $\boldPi$ be the corresponding permutation indicator matrix,
i.e., 
\begin{align*}
\boldPi \in  \{0,1\}^{n \times n},~
\boldPi{\boldone}_n = {\boldone}_n,~
\textnormal{and}~
\boldPi^\top {\boldone}_n = {\boldone}_n,
\end{align*}
where ${\boldone}_n $ is the $n$-dimensional vector with all ones
and $^\top$ denotes the transpose. 
Let us denote the samples matched by a permutation $\pi$ by
\[
Z(\boldPi) := \{(\boldx_i,\boldy_{\pi(i)})\}_{i=1}^n.
\]
The optimal permutation, denoted by $\boldPi^\ast$, can be obtained 
as the maximizer of the dependency between the two sets $\{\boldx_i\}_{i=1}^n$ and $\{\boldy_{i}\}_{i=1}^n$:
\begin{align*}
\boldPi^* := \mathop{\text{argmax}}_{\boldPi} D(Z(\boldPi)),
\end{align*}
where $D$ is some dependence measure.

\section{Existing Methods} \label{sec:related}
In this section, we review two existing methods for CDOM,
and point out their weaknesses.
\subsection{Kernelized Sorting with Mutual Information}
\emph{Kernelized sorting with mutual information} (KS-MI) \citep{COLT:Jebara:2004} matches 
objects in different domains so that MI between matched pairs is maximized.
Here, we review KS-MI following alternative derivation provided in \citet{PAMI:Quadrianto+etal:2010}.

MI is one of the popular dependence measures between random variables. 
For random variables $X$ and $Y$, MI is defined 
as follows \citep{book:Cover+Thomas:2006}:
\[
\mathrm{MI}(Z):=\int p(X,Y)\log\frac{p(X,Y)}{p(X)p(Y)}\mathrm{d}X\mathrm{d}Y,
\]
where $p(X,Y)$ denotes the joint density of $X$ and $Y$,
and $p(X)$ and $p(Y)$ are marginal densities of $X$ and $Y$, respectively.
MI is zero if and only if $X$ and $Y$ are independent,
and thus it may be used as a dependency measure.
Let $H(X)$, $H(Y)$, and $H(X,Y)$ be
the entropies of $X$ and $Y$ and the joint entropy of $X$ and $Y$, respectively:
\begin{align*}
H(X) &= - \int p(X) \log p(X) \mathrm{d}X, \\
H(Y) &= - \int p(Y) \log p(Y) \mathrm{d}Y, \\
H(X, Y) & = - \int p(X, Y) \log p(X,Y) \mathrm{d}X \mathrm{d}Y,
\end{align*}
respectively.
Then the mutual information between $X$ and $Y$ can be written as 
\begin{align*}
\mathrm{MI}(Z) &= H(X) + H(Y) - H(X, Y).
\end{align*}
Since $H(X)$ and $H(Y)$ are independent of permutation $\boldPi$,
maximizing mutual information is equivalent to minimizing the joint entropy $H(X,Y)$.
If $p(X,Y)$ is Gaussian with covariance matrix $\boldSigma$,
the joint entropy is expressed as
\[
 H(X,Y) =  \frac{1}{2} \log |\boldSigma| + \textnormal{Const},
\]
where $|\boldSigma|$ denotes the determinant of matrix $\boldSigma$.

Now, let us assume that $\boldx$ and $\boldy$ are jointly normal
in some reproducing Kernel Hilbert Spaces (RKHSs)
endowed with joint kernel $K(\boldx,\boldx')L(\boldy,\boldy')$, 
where $K(\boldx,\boldx')$ and $L(\boldy,\boldy')$ are reproducing kernels
for $\boldx$ and $\boldy$, respectively.
Then KS-MI is formulated as follows:
\begin{align}
\label{eq:mi}
\min_{\boldPi} \log |\boldGamma 
(\boldK \circ (\boldPi^\top \boldL \boldPi)) \boldGamma|,
\end{align}
where $\boldK = \{K(\boldx_i, \boldx_j)\}_{i,j=1}^n$ and $\boldL = \{L(\boldy_i, \boldy_j)\}_{i,j=1}^n$ are kernel matrices, $\circ$ denotes the Hadamard product (a.k.a.~the element-wise product),
$\boldGamma = \boldI_n - \frac{1}{n}{\boldone}_n{\boldone}_n^\top$ is the centering matrix, and $\boldI_n$ is the $n$-dimensional identity matrix.

A critical weakness of KS-MI is the Gaussian assumption,
which may not be fulfilled in practice.
\subsection{Kernelized Sorting with Hilbert-Schmidt Independence Criterion}
\emph{Kernelized sorting with Hilbert-Schmidt independence criterion} (KS-HSIC)
matches objects in different domains so that
HSIC between matched pairs is maximized.

HSIC is a kernel-based dependence measure given as follows \citep{ALT:Gretton+etal:2005}:
\[
{\textnormal{HSIC}}(Z) = \textnormal{tr}(\bar{\boldK} \bar{\boldL}),
\]
where $\bar{\boldK} = \boldGamma \boldK \boldGamma$
and $\bar{\boldL} = \boldGamma \boldL \boldGamma$ are the centered kernel matrices
for $\boldx$ and  $\boldy$, respectively.
Note that smaller HSIC scores mean that $X$ and $Y$ are closer to be independent.

KS-HSIC is formulated as follows \citep{PAMI:Quadrianto+etal:2010}:
\begin{align}
\max_{\boldPi} {\textnormal{HSIC}}(Z(\boldPi)),\label{eq:maxHSIX}
\end{align}
where 
\begin{align}
{\textnormal{HSIC}}(Z(\boldPi)) = \textnormal{tr}(\bar{\boldK} \boldPi^\top \bar{\boldL} \boldPi).
\label{eq:HSICQAP}
\end{align}
This optimization problem is called  the \emph{quadratic assignment problem} (QAP) \citep{ADM:Finke+etal:1987},
and it is known to be \emph{NP-hard}. There exists several QAP solvers such as
methods based on simulated annealing, tabu search, and genetic algorithms.
However, those QAP solvers are not easy to use in practice since they
contain various tuning parameters.

Another approach to solving Eq.\eqref{eq:maxHSIX}
based on a \emph{linear assignment problem} (LAP) \citep{NRLQ:Kuhn:1955}
was proposed in \citet{PAMI:Quadrianto+etal:2010},
which is explained below.
Let us relax the permutation indicator matrix $\boldPi$ to take real values:
\begin{align}
\boldPi \in [0,1]^{n \times n},~
\boldPi{\boldone}_n = {\boldone}_n,~
\textnormal{and}~
\boldPi^\top {\boldone}_n = {\boldone}_n.
\label{boldPi-relaxed}
\end{align}
Then, Eq.\eqref{eq:HSICQAP} is convex with respect to $\boldPi$
\citep[see Lemma 7 in][]{PAMI:Quadrianto+etal:2010},
and its lower bound can be obtained using some $\widetilde{\boldPi}$ as follows:
\begin{align*}
&\textnormal{tr}(\bar{\boldK} \boldPi^\top \bar{\boldL} \boldPi)\\
 &\geq \textnormal{tr}(\bar{\boldK} \widetilde{\boldPi}^\top \bar{\boldL} \widetilde{\boldPi})
+ \inner{\boldPi - \widetilde{\boldPi}}{\frac{\partial {\textnormal{HSIC}}(Z(\widetilde{\boldPi}))}{\partial \boldPi}} \\
&= 2\textnormal{tr}(\bar{\boldK} \boldPi^\top \bar{\boldL} \widetilde{\boldPi}) - \textnormal{tr}(\bar{\boldK} \widetilde{\boldPi}^\top \bar{\boldL} \widetilde{\boldPi}),
\end{align*}
where $\inner{\cdot}{\cdot}$ denotes the inner product between matrices.
Based on the above lower bound,
\citet{PAMI:Quadrianto+etal:2010} proposed to update the permutation matrix as
\begin{align}
\label{eq:HSICLAP}
\boldPi^{\textnormal{new}} = (1 - \eta)\boldPi^{\textnormal{old}} + \eta \mathop{\text{argmax}}_{\boldPi} \textnormal{tr} \left(\boldPi^\top \bar{\boldL} \boldPi^{\textnormal{old}}  \bar{\boldK} \right),
\end{align}
where $0< \eta \leq 1$ is a step size.
The second term is an LAP subproblem,
which can be efficiently solved by using 
the \emph{Hungarian method}.

In the original KS-HSIC paper \citep{PAMI:Quadrianto+etal:2010},
a C++ implementation of the Hungarian method provided by
Cooper\footnote[1]{http://mit.edu/harold/www/code.html}
was used for solving Eq.\eqref{eq:HSICLAP};
then $\boldPi$ is kept updated by Eq.\eqref{eq:HSICLAP} until convergence.

In this iterative optimization procedure, the choice of initial permutation matrices
is critical to obtain a good solution.
\citet{PAMI:Quadrianto+etal:2010} proposed 
the following initialization scheme.
Suppose the kernel matrices $\bar{\boldK}$ and $\bar{\boldL}$ are rank one,
i.e., for some $\boldf$ and $\boldg$,
$\bar{\boldK}$ and $\bar{\boldL}$ can be expressed as
$\bar{\boldK} = \boldf \boldf^\top$ and $\bar{\boldL} =  \boldg \boldg^\top$.
Then HSIC can be written as
\begin{align}
{\textnormal{HSIC}}(Z(\boldPi)) &= \| \boldf^\top \boldPi \boldg \|^2.
\label{eq:HSICINI}
\end{align}
The initial permutation matrix is determined so that Eq.\eqref{eq:HSICINI} is maximized.
According to Theorems 368 and 369 in \citet{Book:Hardy:1952},
the maximum of Eq.\eqref{eq:HSICINI} is attained
when the elements of $\boldf$ and $\boldPi \boldg$ are ordered in the same way.
That is, if the elements of $\boldf$ are ordered in the ascending manner
(i.e., $f_1 \leq f_2 \leq \cdots \leq f_n$), 
the maximum of Eq.\eqref{eq:HSICINI} is attained
by ordering the elements of $\boldg$ in the same ascending way.
However, since the kernel matrices
$\bar{\boldK}$ and $\bar{\boldL}$ may not be rank one
in practice,
the principal eigenvectors of $\bar{\boldK}$ and $\bar{\boldL}$
were used as $\boldf$ and $\boldg$ in the original KS-HSIC paper
 \citep{PAMI:Quadrianto+etal:2010}.
 We call this \emph{eigenvalue-based initialization}.


Since HSIC is a distribution-free dependence measure,
KS-HSIC is more flexible than KS-MI.
However, a critical weakness of HSIC is that
its performance is sensitive to the choice of kernels
\citep{AAAI:Hal+etal:2010}.
A practical heuristic is to use the Gaussian kernel
with width set to the median distance between samples
\citep[see e.g.,][]{TAS:Fukumizu+etal:2009},
but this does not always work well in practice.

\section{Proposed Methods}
In this section, we propose two alternative CDOM methods that can naturally address
the model selection problem. 

\subsection{Kernelized Sorting with Normalized Cross-Covariance Operator}
The kernel-based dependence measure based on the \emph{normalized cross-covariance operator} (NOCCO) \citep{NIPS:Fukumizu+etal:2008} is given as follows \citep{NIPS:Fukumizu+etal:2008}: 
\[
\textnormal{D}_{\textnormal{NOCCO}}(Z) = \textnormal{tr}(\widetilde{\boldK} \widetilde{\boldL}),
\]
where $\widetilde{\boldK} = \bar{\boldK} (\bar{\boldK} + n \epsilon \boldI_n)^{-1}$,
$\widetilde{\boldL} = \bar{\boldL} (\bar{\boldL} + n \epsilon \boldI_n)^{-1}$,
and $\epsilon > 0$ is a regularization parameter. 
$\textnormal{D}_{\textnormal{NOCCO}}$ was shown to be asymptotically independent of the choice of kernels.
Thus, KS with $\textnormal{D}_{\textnormal{NOCCO}}$ (KS-NOCCO) is expected to be less sensitive to the kernel parameter choice
than KS-HSIC.

The permuted version of $\widetilde{\boldL}$ can be written as
\begin{align*}
\widetilde{\boldL}(\boldPi) &= \boldPi^\top \bar{\boldL} \boldPi (\boldPi^\top \bar{\boldL} \boldPi + n \epsilon \boldI_n)^{-1}\\
                &= \boldPi^\top \bar{\boldL} (\bar{\boldL}  + n \epsilon \boldI_n)^{-1}\boldPi\\
                &= \boldPi^\top \widetilde{\boldL} \boldPi,
\end{align*}
where we used the orthogonality of $\boldPi$ (i.e., $\boldPi^\top \boldPi = \boldPi \boldPi^\top = \boldI_n$). Thus, the dependency measure for $Z(\boldPi)$ can be written as
\begin{align*}
\textnormal{D}_{\textnormal{NOCCO}}(Z(\boldPi)) = \textnormal{tr}(\widetilde{\boldK} \boldPi^\top \widetilde{\boldL} \boldPi).
\end{align*}

Since this is essentially the same form as HSIC,
a local optimal solution may be obtained in the same way as KS-HSIC:
\begin{align}
\label{eq:NOCCOLAP}
\boldPi^{\textnormal{new}} = (1 - \eta)\boldPi^{\textnormal{old}} + \eta \mathop{\text{argmax}}_{\boldPi} \textnormal{tr} \left(\boldPi^\top \widetilde{\boldL} \boldPi^{\textnormal{old}}  \widetilde{\boldK} \right).
\end{align}
However, the property that $\textnormal{D}_{\textnormal{NOCCO}}$ is independent of the kernel choice
holds only asymptotically.
Thus, with finite samples, $\textnormal{D}_{\textnormal{NOCCO}}$ does depend on the choice of kernels
as well as the regularization parameter $\epsilon$
which needs to be manually tuned.

\begin{figure*}[t]
  \centering
\subfigure[Unpaired data]{
\includegraphics[width=0.3\textwidth]{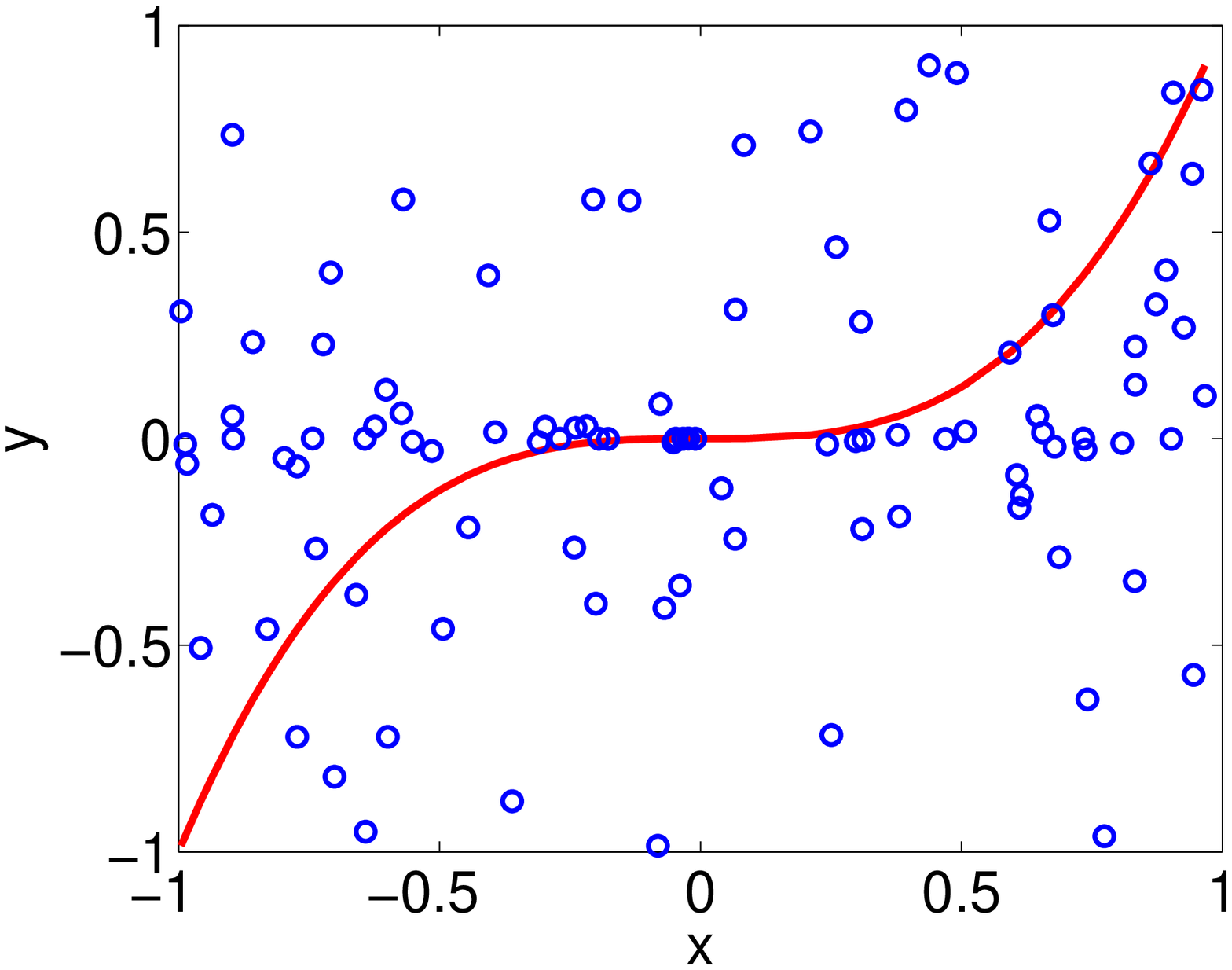}
\label{fig:toy_data}
}
\subfigure[Eigenvalue-based initialization.]{
\includegraphics[width=0.3\textwidth]{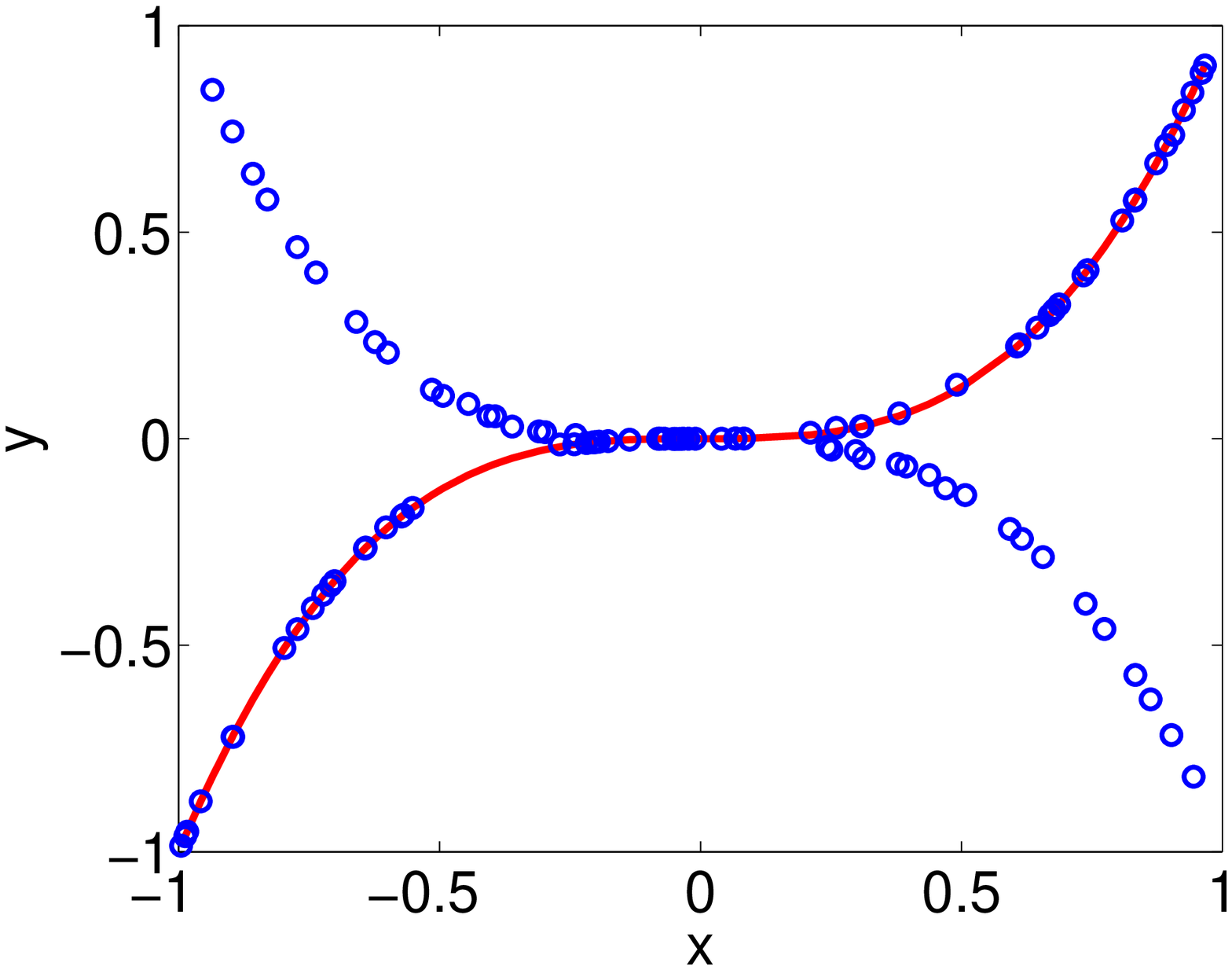}
  \label{fig:toy_itr0}
}
\subfigure[Matched result by KS-NOCCO.]{
\includegraphics[width=0.3\textwidth]{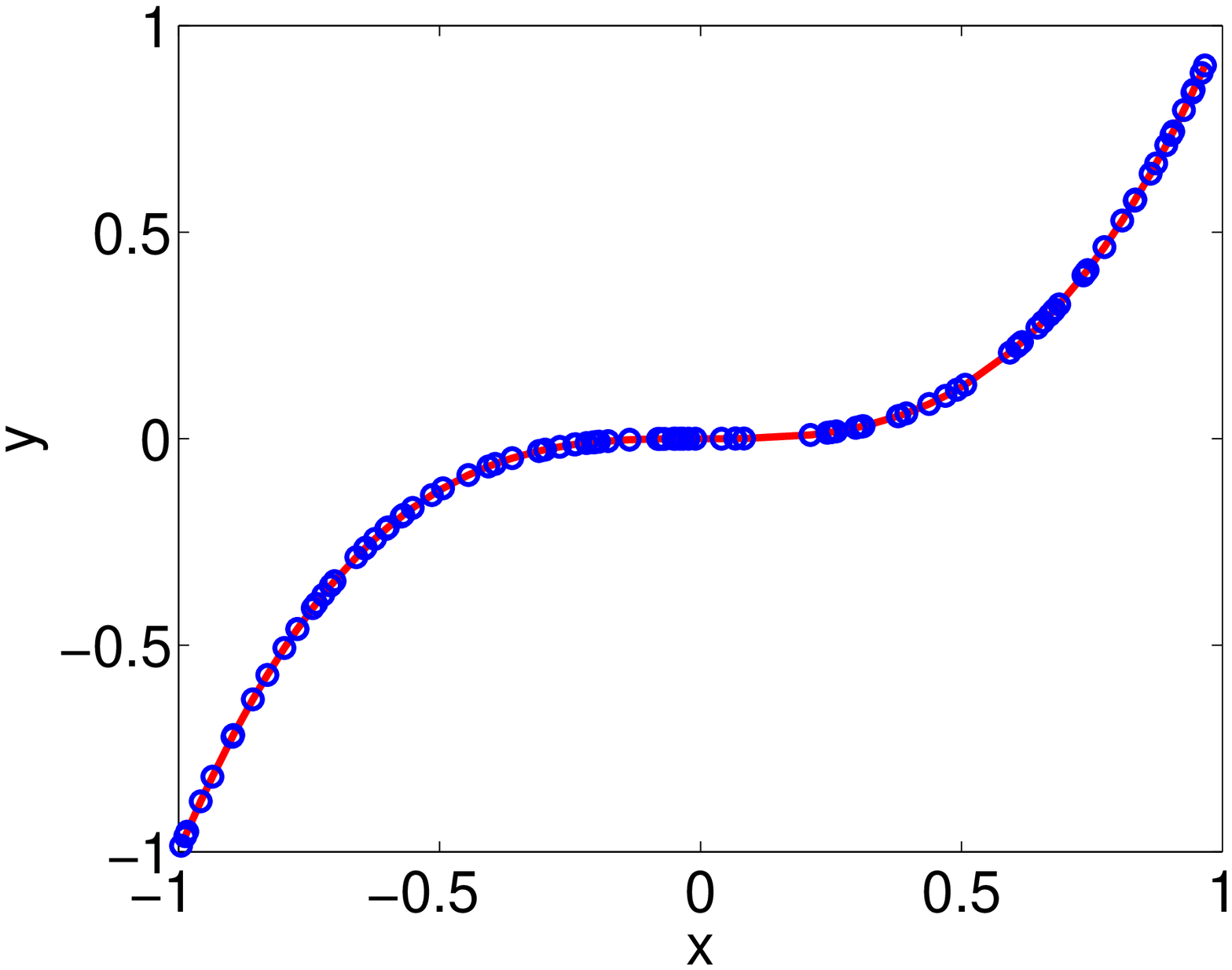}
\label{fig:toy_nocco_itr1}
}
\subfigure[Matched result by LSOM.]{
\includegraphics[width=0.3\textwidth]{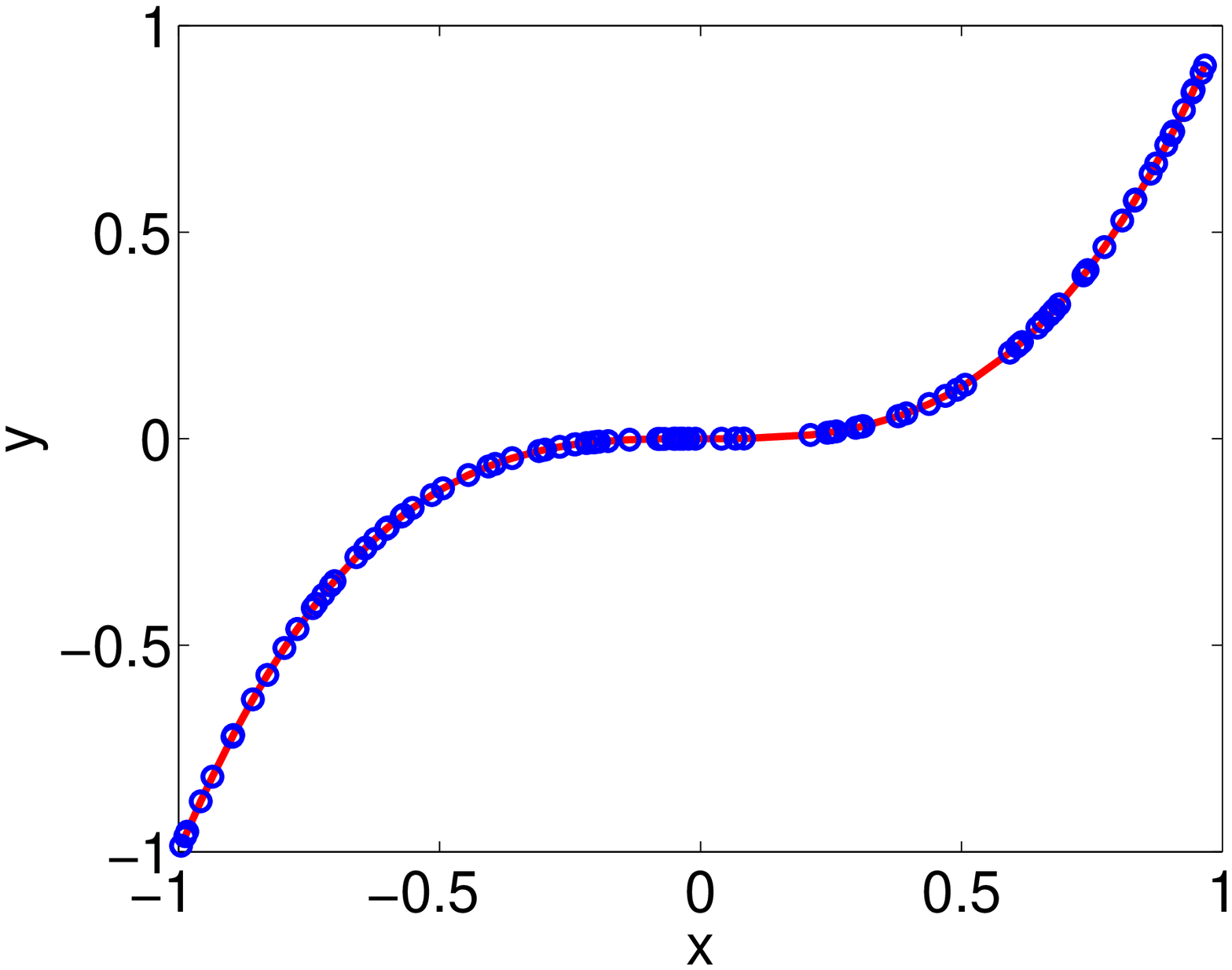}
\label{fig:toy_lsom_itr1}
}
\subfigure[Values of empirical $\textnormal{D}_{\textnormal{NOCCO}}$ score in KS-NOCCO.]{
\includegraphics[width=0.3\textwidth]{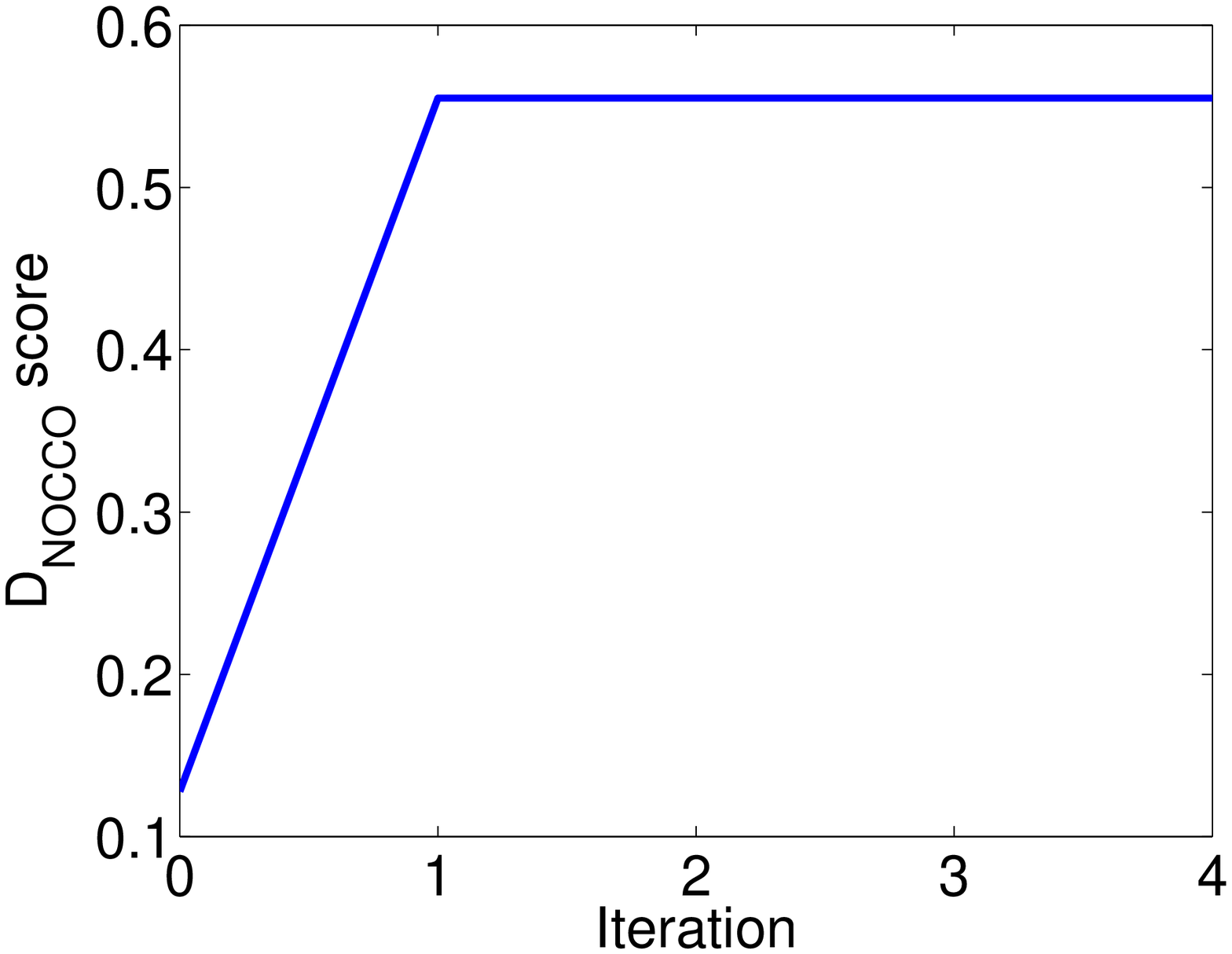}
\label{fig:converge_nocco}
}
\subfigure[Values of empirical SMI score in LSOM.]{
\includegraphics[width=0.3\textwidth]{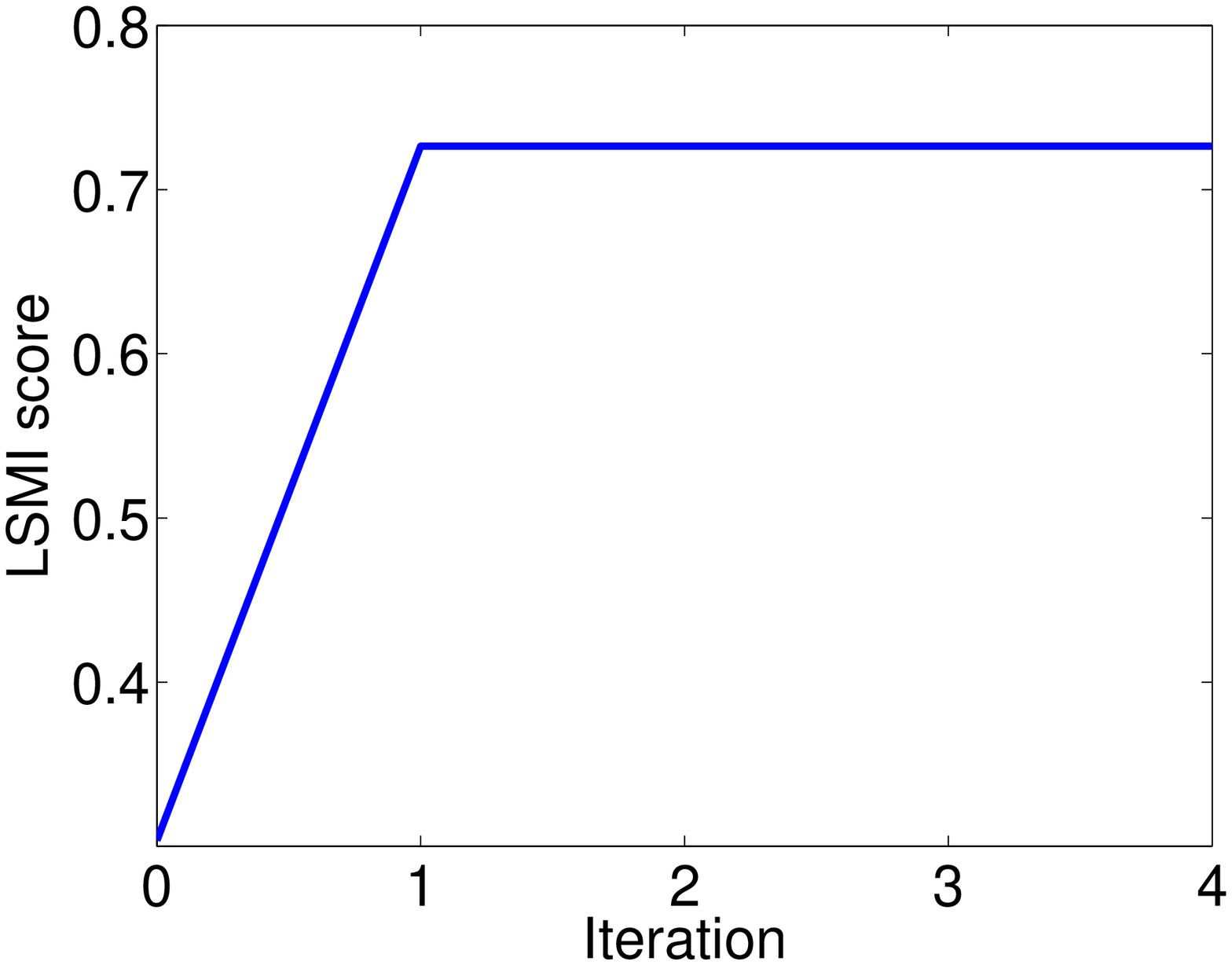}
\label{fig:converge_lsom}
}
 \caption{Illustrative example. (a)--(d):The solid line denotes the
true function and the circles denote samples. (e): Values of empirical $\textnormal{D}_{\textnormal{NOCCO}}$ score as a function of the number of iterations. (f): Values of empirical SMI score as a function of the number of iterations.  }
    \label{fig:toy_data_all}
  \end{figure*}

\begin{figure*}[t]
  \centering
\subfigure[KS-HSIC with different Gaussian kernel widths.]{
\includegraphics[width=0.3\textwidth]{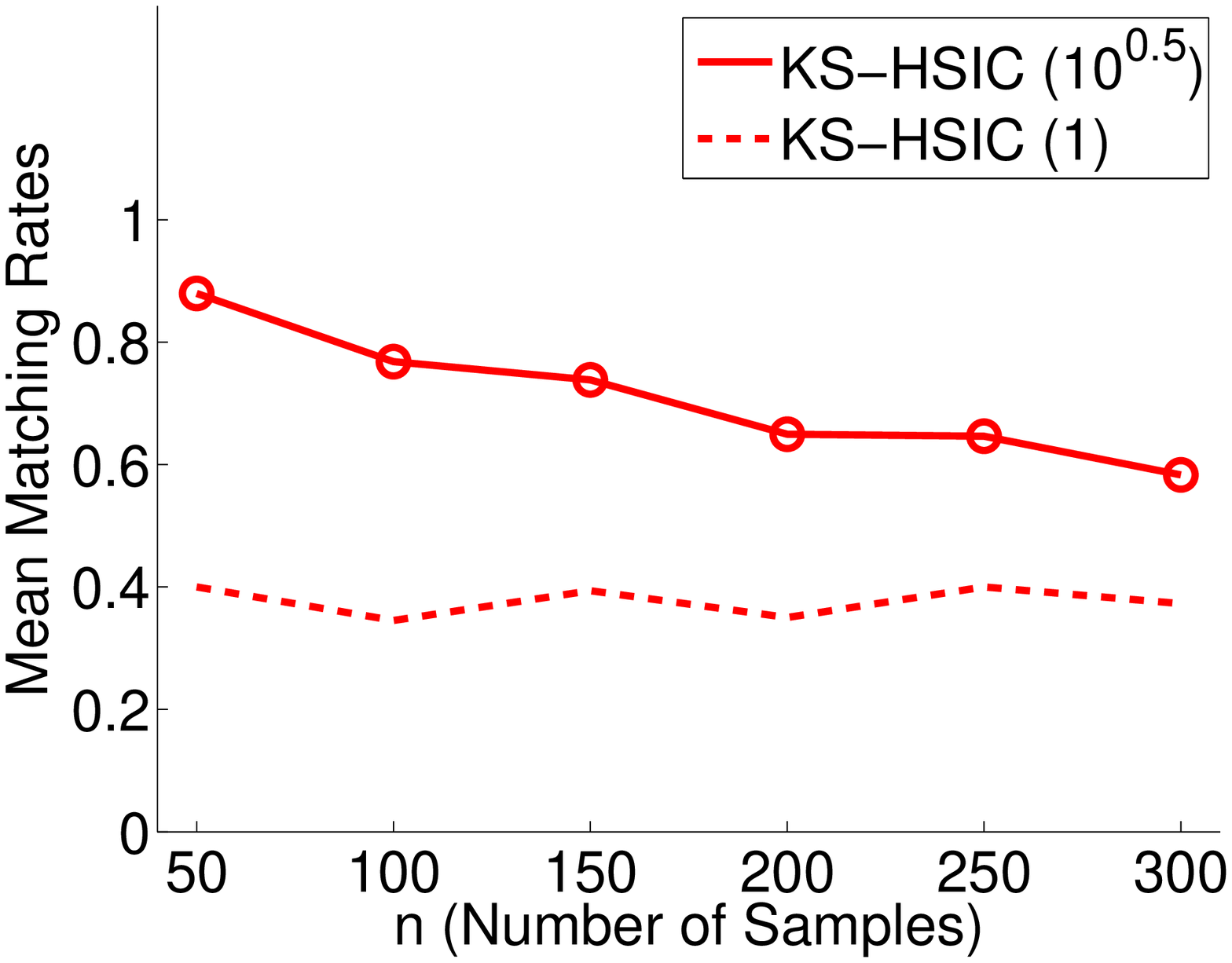}
\label{fig:hsic_image_match}
}
\hspace*{2mm}
\subfigure[KS-NOCCO with different Gaussian kernel widths and regularization parameter.]{
\includegraphics[width=0.3\textwidth]{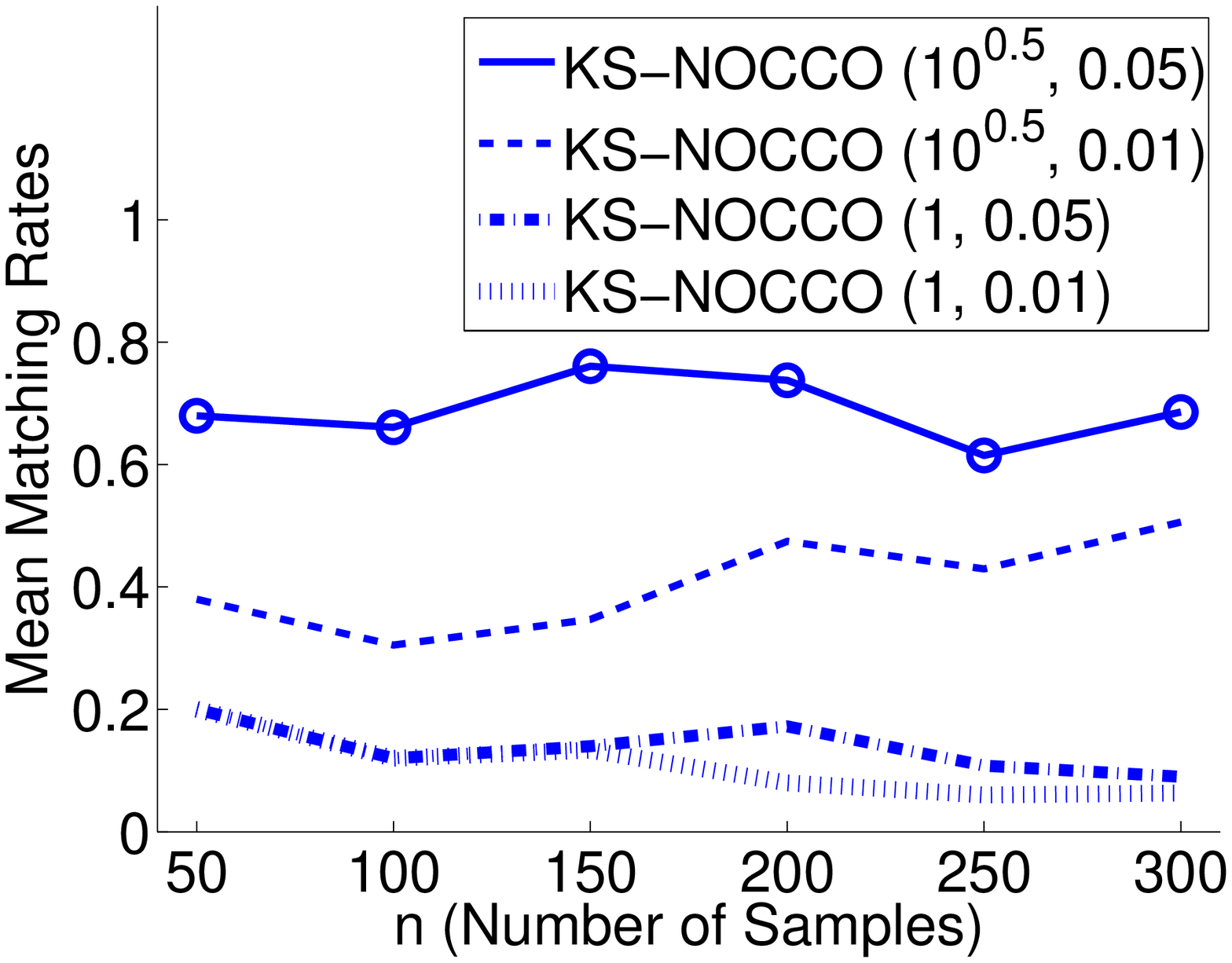}
  \label{fig:nocco_image_match}
}
\hspace*{2mm}
\subfigure[LSOM, optimally-tuned KS-NOCCO, and optimally-tuned KS-HSIC.]{
\includegraphics[width=0.3\textwidth]{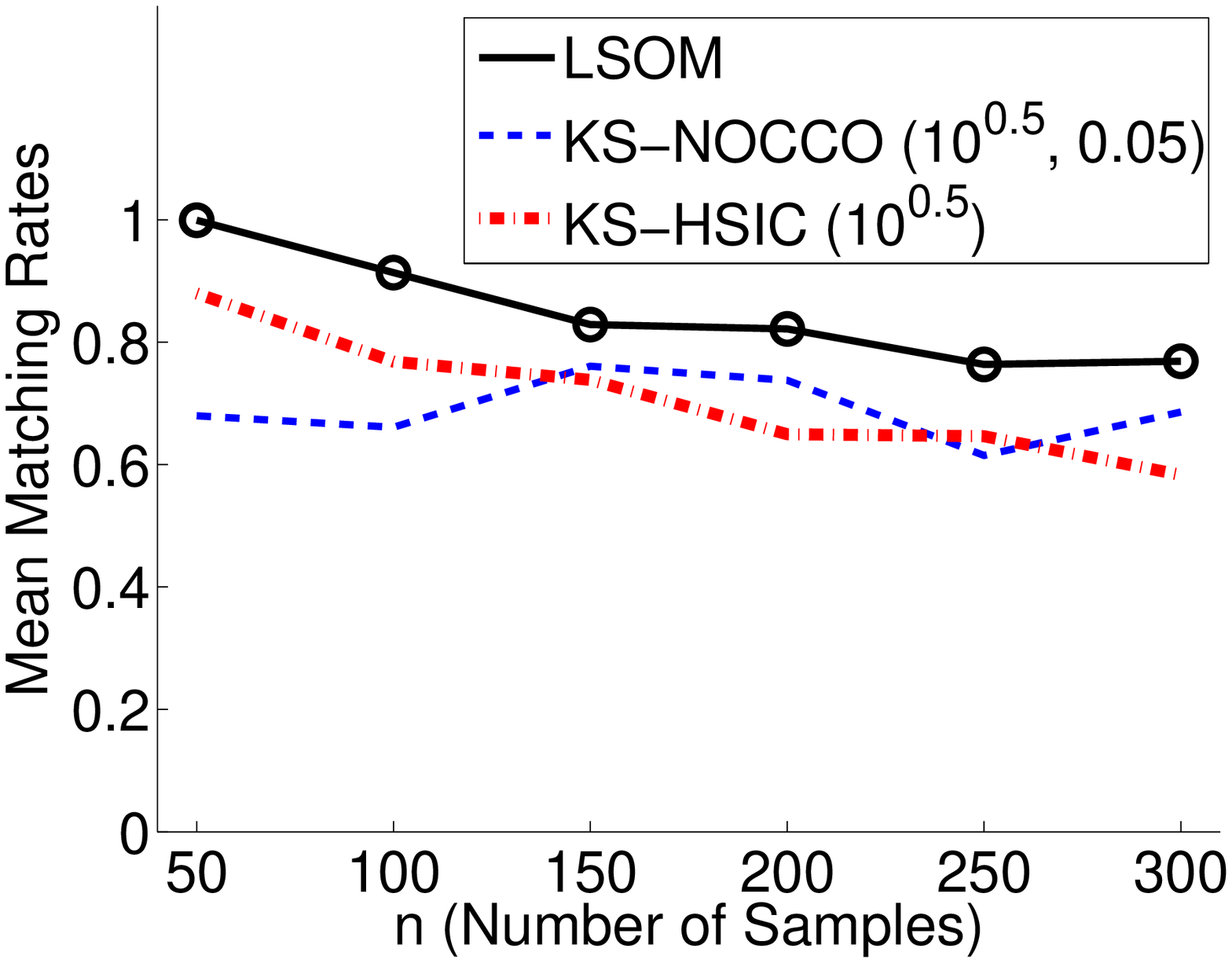}
  \label{fig:lsom_image_match}
}
 \caption{Image matching results. 
The best method in terms of the mean error and comparable methods according to the t-test at the significance level 1\% are specified by `$\circ$'.
}
    \label{fig:image_result}
\end{figure*}
\subsection{Least-Squares Object Matching}
Next, we propose an alternative method called \emph{least-squares object matching} (LSOM),
in which we employ \emph{least-squares mutual information} (LSMI) \citep{BMCBio:Suzuki+etal:2009a}
as a dependency measure.
LSMI is a consistent estimator of the \emph{squared-loss mutual information} (SMI)
achieving the optimal convergence rate. 
SMI is defined and expressed as
\begin{align}
&\textnormal{SMI}(Z)  \nonumber \\
&= \frac{1}{2}\int\int \left(\frac{p(X,Y)}{p(X)p(Y)} - 1\right)^2
p(X)p(Y)\textnormal{d}X \textnormal{d}Y
 \nonumber\\
&= \frac{1}{2}\iint \left(\frac{p(X,Y)}{p(X)p(Y)}\right)p(X,Y)
\textnormal{d}X\textnormal{d}Y - \frac{1}{2}.
\label{eq:SMIdef}
\end{align}
Note that SMI is the \emph{Pearson divergence}
\citep{PhMag:Pearson:1900} from $p(X,Y)$ to $p(X)p(Y)$,
while ordinary MI is the \emph{Kullback-Leibler divergence}
\citep{Annals-Math-Stat:Kullback+Leibler:1951} from $p(X,Y)$ to $p(X)p(Y)$.
SMI is zero if and only if $X$ and $Y$ are independent, as ordinary MI.
Its estimator LSMI is given as follows \citep{BMCBio:Suzuki+etal:2009a}:
\begin{align*}
\mathrm{LSMI}(Z)&= \frac{1}{2}{\boldalpha}^\top{\boldh}   - \frac{1}{2},
\end{align*}
where
\begin{align*}
  {\boldalpha}&= \boldH^{-1} \boldh,\\
  \boldH &= \frac{1}{n^2} \left((\boldK \boldK^\top) \circ (\boldL \boldL^\top) + \lambda\boldI_n\right), \\
  {\boldh}&=
  \left(\frac{1}{n}\boldK \circ \boldL\right) {\boldone}_n.
\end{align*}
Here, $\lambda$ ($\ge0$) is the regularization parameter.
Since cross-validation (CV) with respect to SMI is possible for model selection,
tuning parameters in LSMI (i.e., the Gaussian kernel width and the regularization parameter)
can be objectively optimized. This is a notable advantage over kernel-based approaches. 

Below, we use the following equivalent expression of LSMI:
\begin{align}
{\textnormal{LSMI}}(Z)
     &= \frac{1}{2n} \textnormal{tr} \left( \boldL {\boldA} \boldK \right) - \frac{1}{2},
\label{eq:empSMI-2}
\end{align}
where ${\boldA}$ is the diagonal matrix with diagonal elements
given by $\boldalpha$.
Note that we used Eq.(73) and Eq.(75) in \citet{Tech:Minka:2000}
for obtaining the above expression.

LSMI for the permuted data $Z(\boldPi)$ is given by
\begin{align*}
\mathrm{LSMI}(Z(\boldPi))&=
\frac{1}{2n} \textnormal{tr} \left(\boldPi^\top \boldL \boldPi \boldA_\boldPi \boldK \right) - \frac{1}{2},
\end{align*}
where $\boldA_\boldPi$ is the diagonal matrix with diagonal elements
given by $\boldalpha_\boldPi$,
and $\boldalpha_\boldPi$ is given by
\begin{align*}
  {\boldalpha}_\boldPi&= \boldH_\boldPi^{-1} \boldh_\boldPi,\\
  \boldH_\boldPi &= \frac{1}{n^2} \left((\boldK \boldK^\top) \circ (\boldPi^\top \boldL \boldL^\top \boldPi) + \lambda\boldI_n\right), \\
  {\boldh}_\boldPi &=
  \left(\frac{1}{n}\boldK \circ (\boldPi^\top \boldL \boldPi)\right) {\boldone}_n.
\end{align*}
Consequently, LSOM is formulated as follows:
\begin{align*}
\max_{\boldPi}~ \mathrm{LSMI}(Z(\boldPi)).
\end{align*}
Since this optimization problem is in general NP-hard and is not convex,
we simply use the same optimization strategy as KS-HSIC,
i.e., for the current $\boldPi^{\textnormal{old}}$, the solution is updated as
\begin{align}
\label{eq:LSOMLAP}
\boldPi^{\textnormal{new}} &= 
 (1 - \eta)\boldPi^{\textnormal{old}} + \eta 
\mathop{\text{argmax}}_{\boldPi}~ \textnormal{tr} \left(\boldPi^\top \boldL \boldPi^{\textnormal{old}} \boldA_{\boldPi^{\textnormal{old}}} \boldK \right).
\end{align}


\section{Experiments} \label{sec:experiments}
In this section, we first illustrate the behavior of the proposed methods
using a toy data set,
and then experimentally evaluate our proposed algorithms in the image matching, unpaired voice conversion, and photo album summarization tasks.

In all the methods, we use the Gaussian kernels:
\begin{align*}
K(\boldx, \boldx') &= \exp \left(-\frac{\|\boldx - \boldx'\|^2}{2\sigma_{\mathrm{x}}^2} \right), \\
L(\boldy, \boldy') &= \exp \left(-\frac{\|\boldy - \boldy'\|^2}{2\sigma_{\mathrm{y}}^2} \right),
\end{align*}
and we set the maximum number of iterations
for updating permutation matrices to 20 and the step size $\eta$ to 1.
To avoid falling into undesirable local optima,
optimization is carried out 10 times with different initial permutation matrices,
which are determined by the eigenvalue-based initialization heuristic
with Gaussian kernel widths
\begin{align*}
  (\sigma_\mathrm{x}, \sigma_\mathrm{y})=c\times(m_\mathrm{x}, m_\mathrm{y}),
\end{align*}
where $c =1^{1/2}, 2^{1/2}, \ldots, 10^{1/2}$, and
\begin{align*}
  m_\mathrm{x} &= 2^{-1/2}\textnormal{median}(\{\|\boldx_i - \boldx_j\|\}_{i,j=1}^n),\\
m_\mathrm{y} &=  2^{-1/2}\textnormal{median}(\{\|\boldy_i - \boldy_j\|\}_{i,j=1}^n).
\end{align*}

In KS-HSIC and KS-NOCCO, we use the Gaussian kernel with
the following widths:
\begin{align*}
  (\sigma_\mathrm{x}, \sigma_\mathrm{y})=c'\times(m_\mathrm{x}, m_\mathrm{y}),
\end{align*}
where $c'=1^{1/2}, 10^{1/2}$.
In KS-NOCCO, we use the following regularization parameters:
\begin{align*}
\epsilon =0.01, 0.05.
\end{align*}
In LSOM, we choose the model parameters of LSMI,
$\sigma_\mathrm{x}$, $\sigma_\mathrm{y}$, and $\lambda$ by 2-fold CV from
\begin{align*}
(\sigma_\mathrm{x}, \sigma_\mathrm{y}) &= c\times(m_\mathrm{x}, m_\mathrm{y}), \\
\lambda &=10^{-1}, 10^{-2}, 10^{-3}.
\end{align*}

\subsection{Illustrative Example}\label{sec:illusrate}
Here, we illustrate the behavior of the proposed KS-NOCCO and LSOM using a toy matching dataset. 

Let us consider the following regression model:
\[
Y = X^3,
\]
where $X$ is subject to the uniform distribution on $(-1, 1)$. We draw 100 paired samples of $X$ and $Y$ following the above generative model (i.e, $\{(x_i, y_i)\}_{i=1}^{100}$). 
Then, given that $\{y_i\}_{i=1}^{100}$ are randomly shuffled, the goal is to recover the original correspondence. In KS-NOCCO, we set the Gaussian kernel width to
\begin{align*}
  (\sigma_\mathrm{x}, \sigma_\mathrm{y})=10^{1/2}\times(m_\mathrm{x}, m_\mathrm{y}),
\end{align*}
and $\epsilon = 0.05$.

Figure~\ref{fig:toy_data} shows the original unpaired data, where the true function is shown by the solid line. Figure~\ref{fig:toy_itr0} shows the matched pairs with eigenvalue-based initialization, and Figures~\ref{fig:toy_nocco_itr1} and \ref{fig:toy_lsom_itr1} show the matched pairs by KS-NOCCO and LSOM.
The graphs show that matching are performed correctly by KS-NOCCO and LSOM. 
Figures~\ref{fig:converge_nocco} and \ref{fig:converge_lsom} show the values of $\textnormal{D}_{\textnormal{NOCCO}}$ and LSMI scores as functions of the number of iterations.
This shows that a local optimal solution has been obtained only in one iteration.
\subsection{Image Matching}
Next, let us consider a toy image matching problem:
we vertically divide images of size $40 \times 40$ pixels in the middle, and
make two sets of half-images $\{\boldx_i\}_{i=1}^{n}$ and $\{\boldy_i\}_{i=1}^{n}$.
Given that $\{\boldy_i\}_{i=1}^{n}$ is randomly permuted, the goal is to recover the correct correspondence.

\begin{figure}[t]
  \centering
  \includegraphics[width=0.45\textwidth]{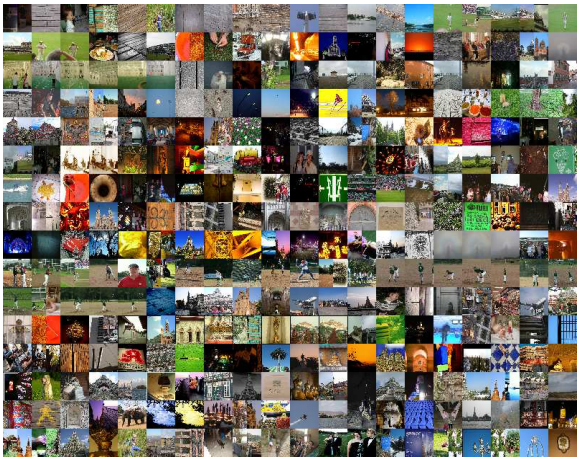}
  \caption{Image matching result by LSOM. In this case,
    234 out of 320 images (73.1\%) are matched correctly.}
  \label{fig:image_match}
\vspace{-0.5cm}
\end{figure}

Figure~\ref{fig:image_result} summarizes the average correct matching rate
over $100$ runs as functions of the number of images,
showing that the proposed LSOM method tends to outperform
the best tuned KS-NOCCO and KS-NOCCO methods.
Figure~\ref{fig:image_match} depicts an example of image matching results
obtained by LSOM, showing that most of the images are matched correctly.

\begin{figure*}[t]
  \centering
\subfigure{
\includegraphics[width=0.3\textwidth]{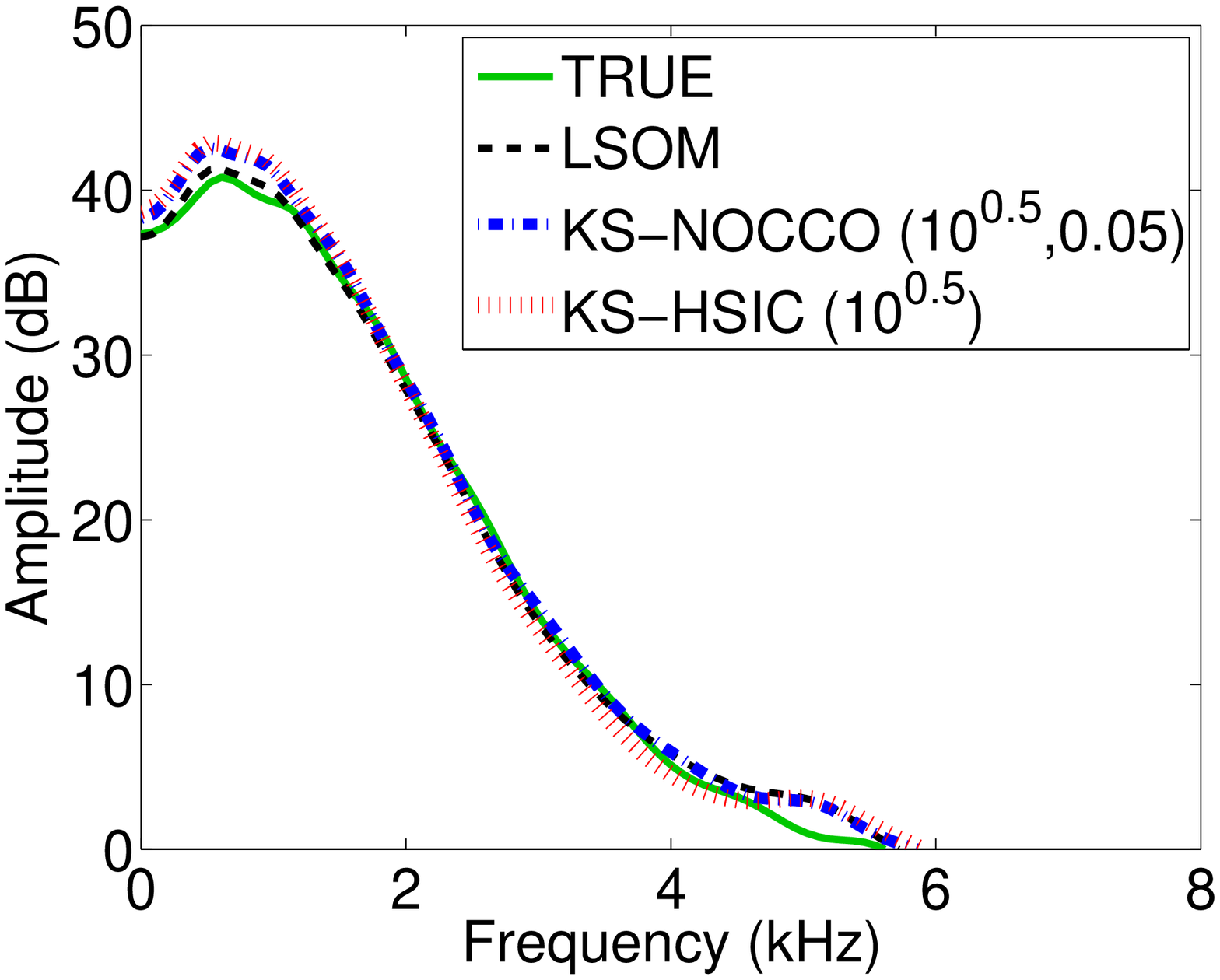}
\label{fig:senv_1}
}
\subfigure{
\includegraphics[width=0.3\textwidth]{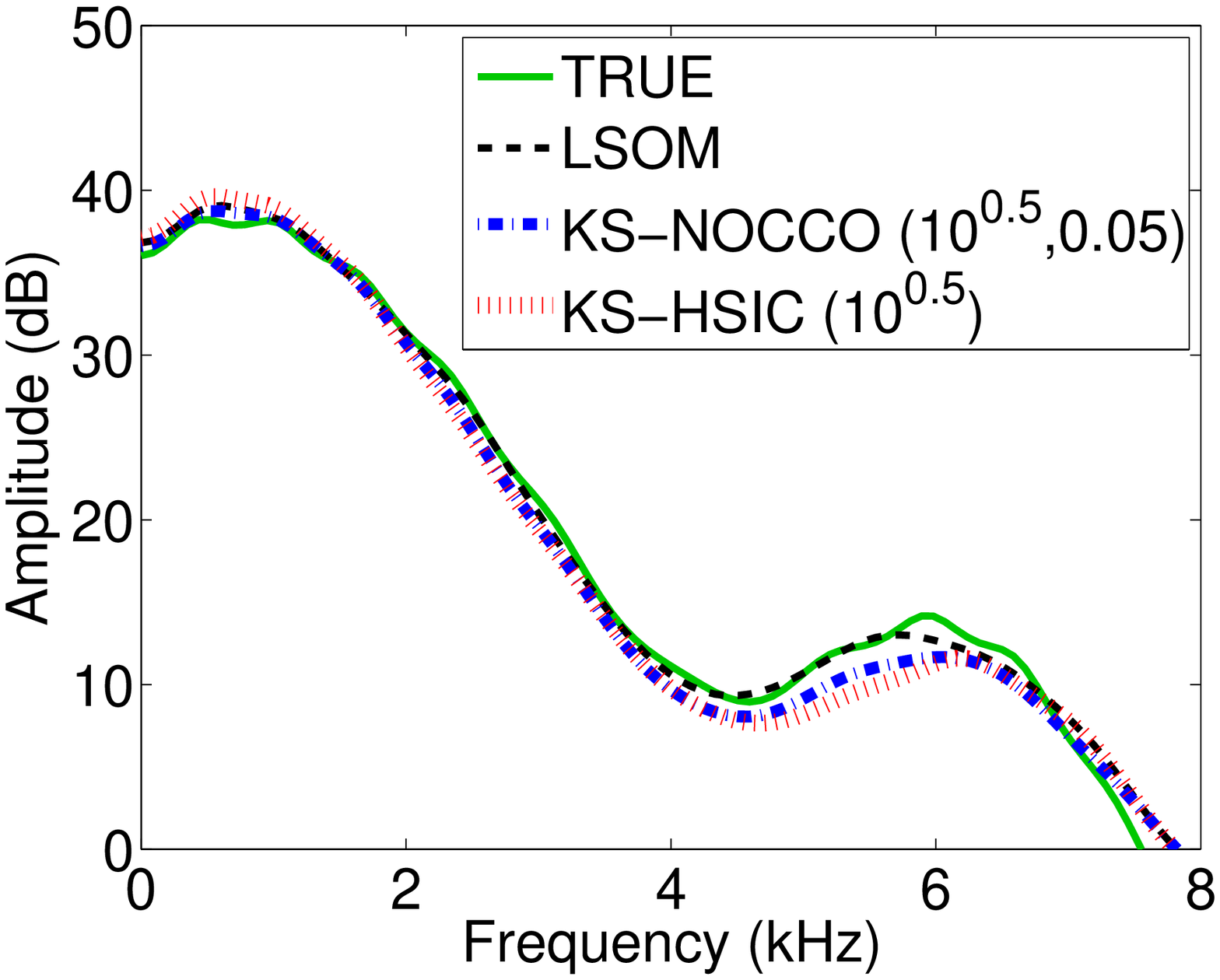}
  \label{fig:senv_2}
}
\subfigure{
\includegraphics[width=0.3\textwidth]{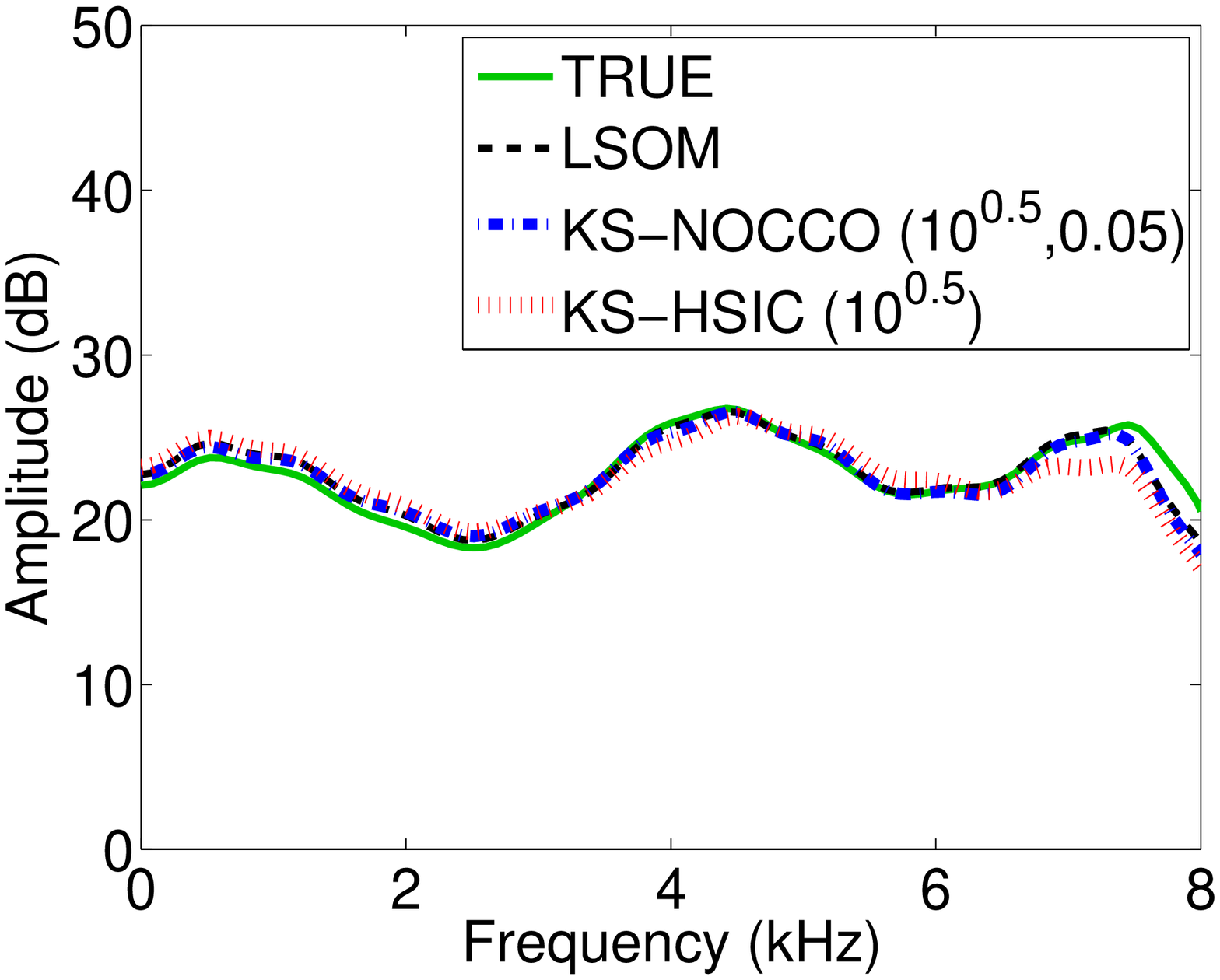}
  \label{fig:senv_3}
}
 \caption{True spectral envelopes and their estimates.
 }
    \label{fig:senv_result}
\end{figure*}

\subsection{Unpaired Voice Conversion}
Next, we consider an unpaired voice conversion task,
which is aimed at matching the voice of a source speaker with that of a target speaker.

In this experiment, we use 200 short utterance samples recorded from
two male speakers in French, with sampling rate 44.1kHz.  We first
convert the utterance samples to 50-dimensional \emph{line spectral frequencies}
(LSF) vector \citep{ICASSP:Kain+etal:1998}. We denote the source and target LSF
vectors by $\boldx$ and $\boldy$, respectively.
Then the voice conversion task can be regarded as a multi-dimensional regression problem
of learning a function from $\boldx$ to $\boldy$.
However, different from a standard regression setup,
paired training samples are not available; instead,
only unpaired samples $\{\boldx_i\}_{i=1}^{n}$ and $\{\boldy_i\}_{i=1}^{n}$ are given.

By CDOM, we first match $\{\boldx_i\}_{i=1}^{n}$ and $\{\boldy_i\}_{i=1}^{n}$,
and then we train a multi-dimensional kernel regression model \citep{book:Schoelkopf+Smola:2002}
using the matched samples $\{(\boldx_{\pi(i)},\boldy_i)\}_{i=1}^{n}$ as 
\begin{align*}
\min_{\boldW} \sum_{i = 1}^{n} \|\boldy_i - \boldW^\top \boldk ({\boldx}_{\pi(i)})\|^2 + \frac{\delta}{2}\textnormal{tr}(\boldW^\top \boldW),
\end{align*}
where
\begin{align*}
{\bm k}({\boldx}) &=
 (K({\boldx},{\boldx}_{{\pi}(1)}), \ldots,
 K({\boldx},{\boldx}_{{\pi}(n)}))^\top,\\
K(\boldx,\boldx') &= \exp \left(-\frac{\|\boldx - \boldx'\|^2}{2\tau^2} \right).
\end{align*}
Here, $\tau$ is a Gaussian kernel width and $\delta$ is a regularization parameter;
they are chosen by 2-fold CV. %

\begin{figure}[t]
  \centering
  \includegraphics[width=0.45\textwidth]{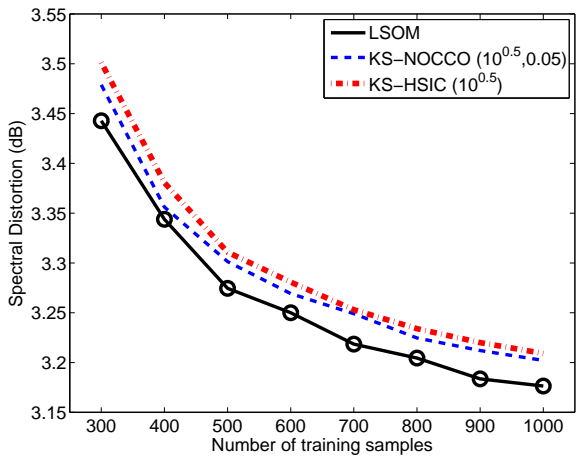}
  \caption{Unpaired voice conversion results. The best method in terms of the mean error and comparable methods according to the t-test at the significance level 1\% are specified by `$\circ$'.
}
  \label{fig:lsom_vc}
\end{figure}

We repeat the experiments 100 times by randomly shuffling training and test samples,
and evaluate the voice convergence performance by 
\emph{log-spectral distance} for $8000$ test samples\footnote{
The smaller the spectral distortion is, the better the quality of voice conversion is.}
\citep{book:Quackenbush+etal:1988}.
 Figure~\ref{fig:senv_result} shows the true spectral envelope and their estimates,
and Figure~\ref{fig:lsom_vc} shows the average performance over 100 runs as the number of training samples.  
These results show that the proposed LSOM tends to outperform KS-NOCCO and KS-HSIC.

\begin{figure*}[t]
  \centering
\subfigure[Layout of 320 images into a 2D grid of size 16 by 20 using LSOM.]{
\includegraphics[width=0.3\textwidth]{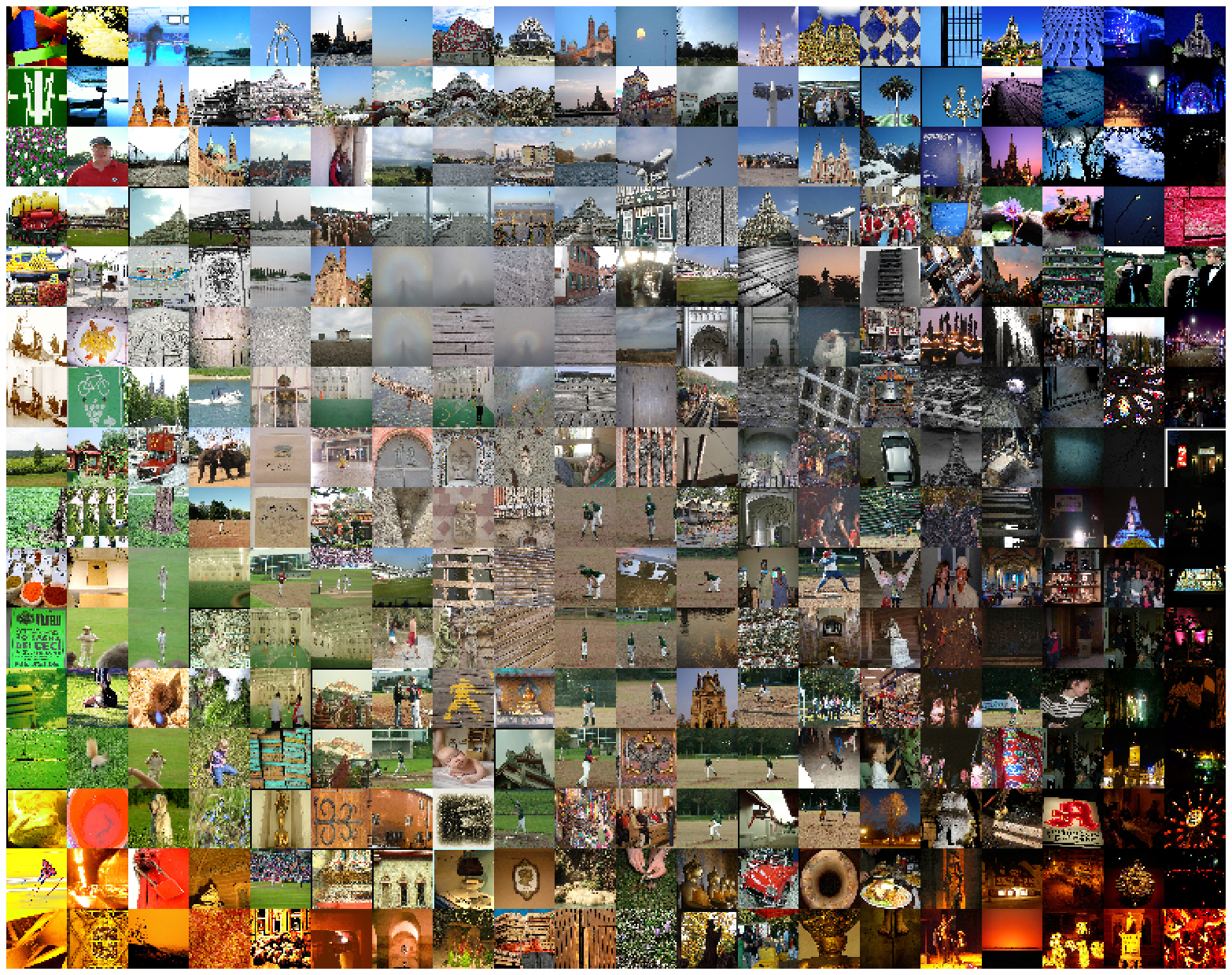}
\label{fig:image_grid}
}
\hspace*{2mm}
\subfigure[Layout of 225 facial images into a 2D grid of size 15 by 15 using LSOM.]{
\includegraphics[width=0.3\textwidth]{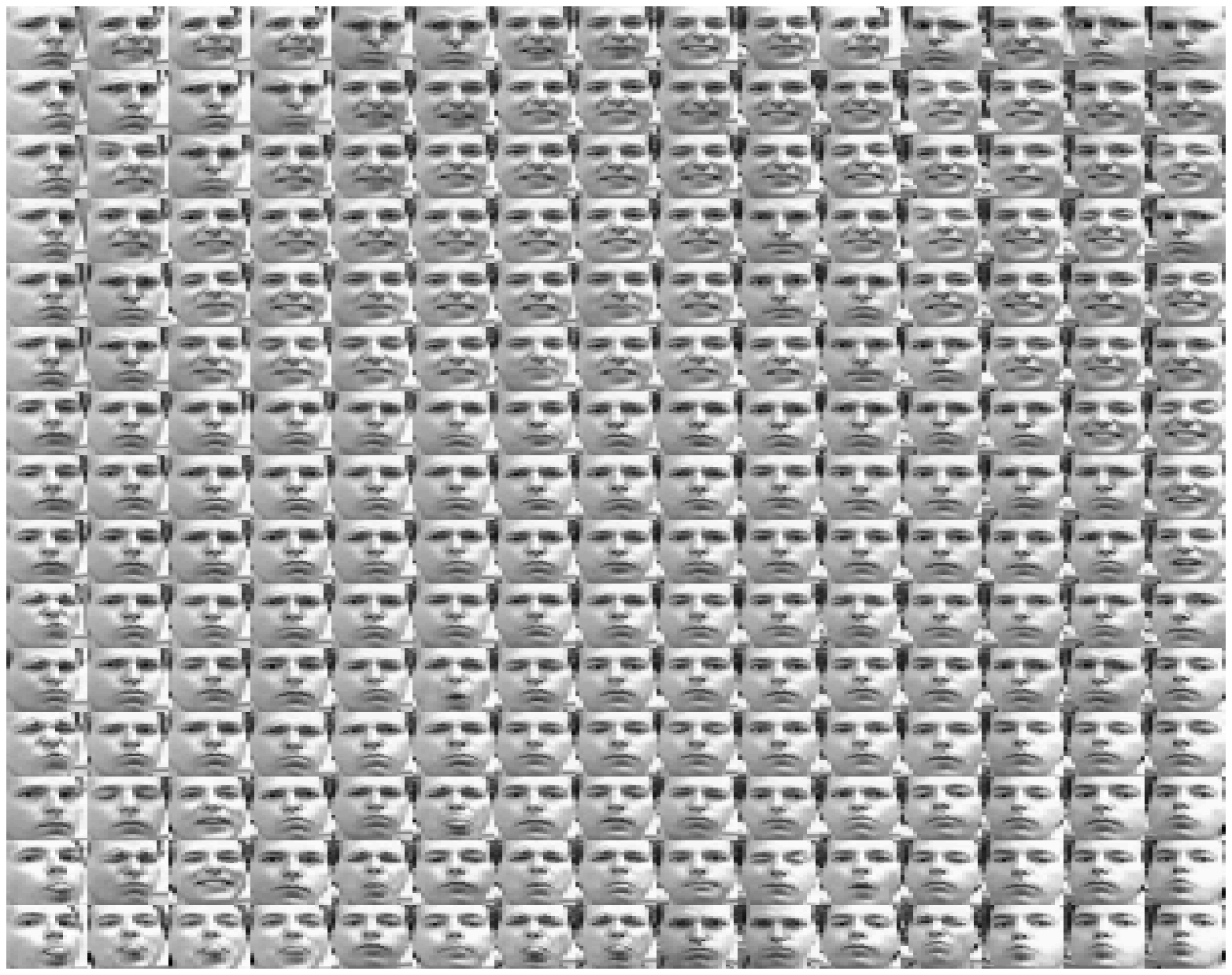}
  \label{fig:image_face}
}
\hspace*{2mm}
\subfigure[Layout of 320 digit `7' into a 2D grid of size 16 by 20 using LSOM.]{
\includegraphics[width=0.3\textwidth]{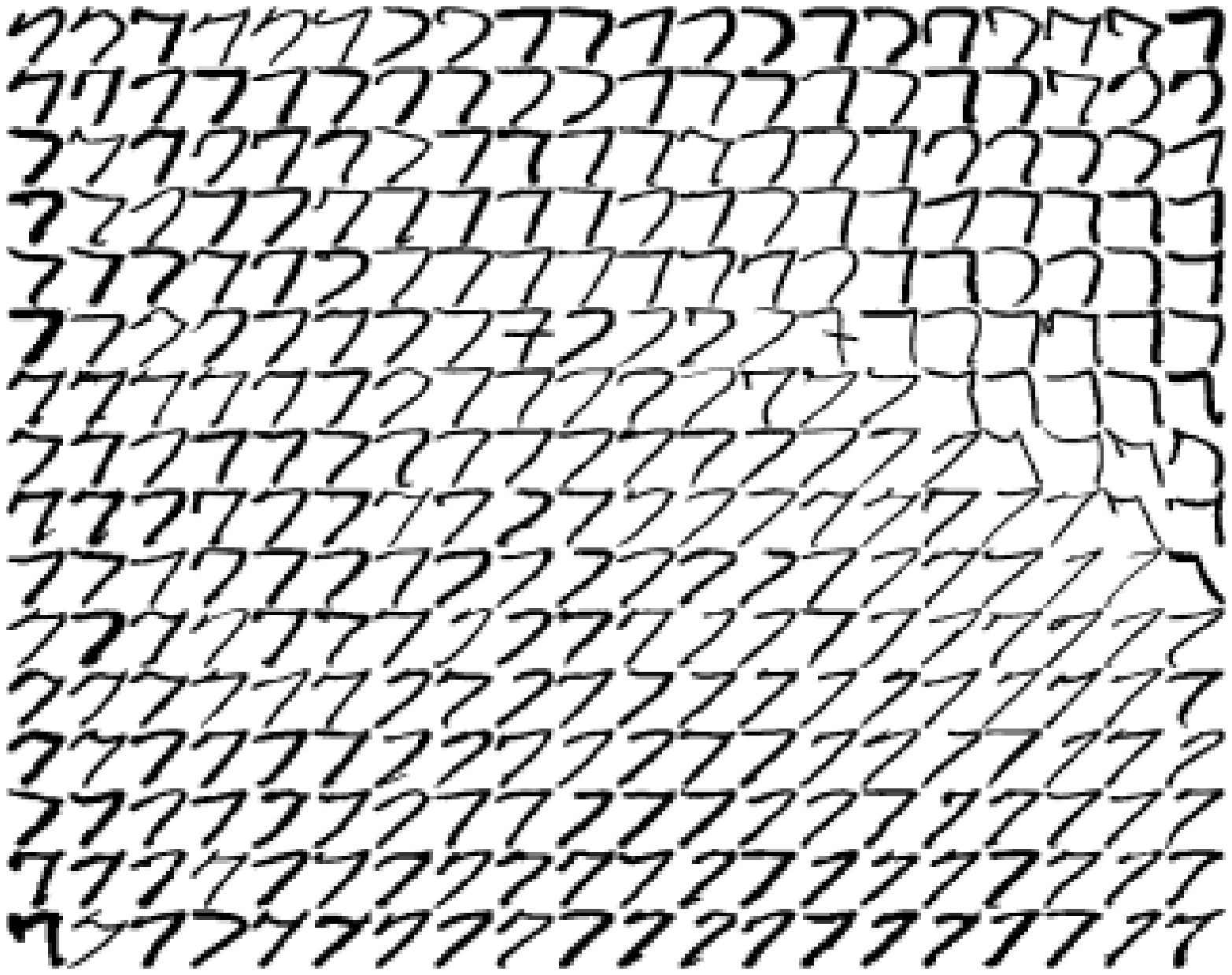}
  \label{fig:usps}
}
 \caption{Images are automatically aligned into rectangular grid frames expressed
in the Cartesian coordinate system.}
    \label{fig:recimages}
\end{figure*}

\begin{figure*}[t]
  \centering
\subfigure[Layout of 120 images into a Japanese character `mountain' by LSOM.]{
\includegraphics[width=0.3\textwidth]{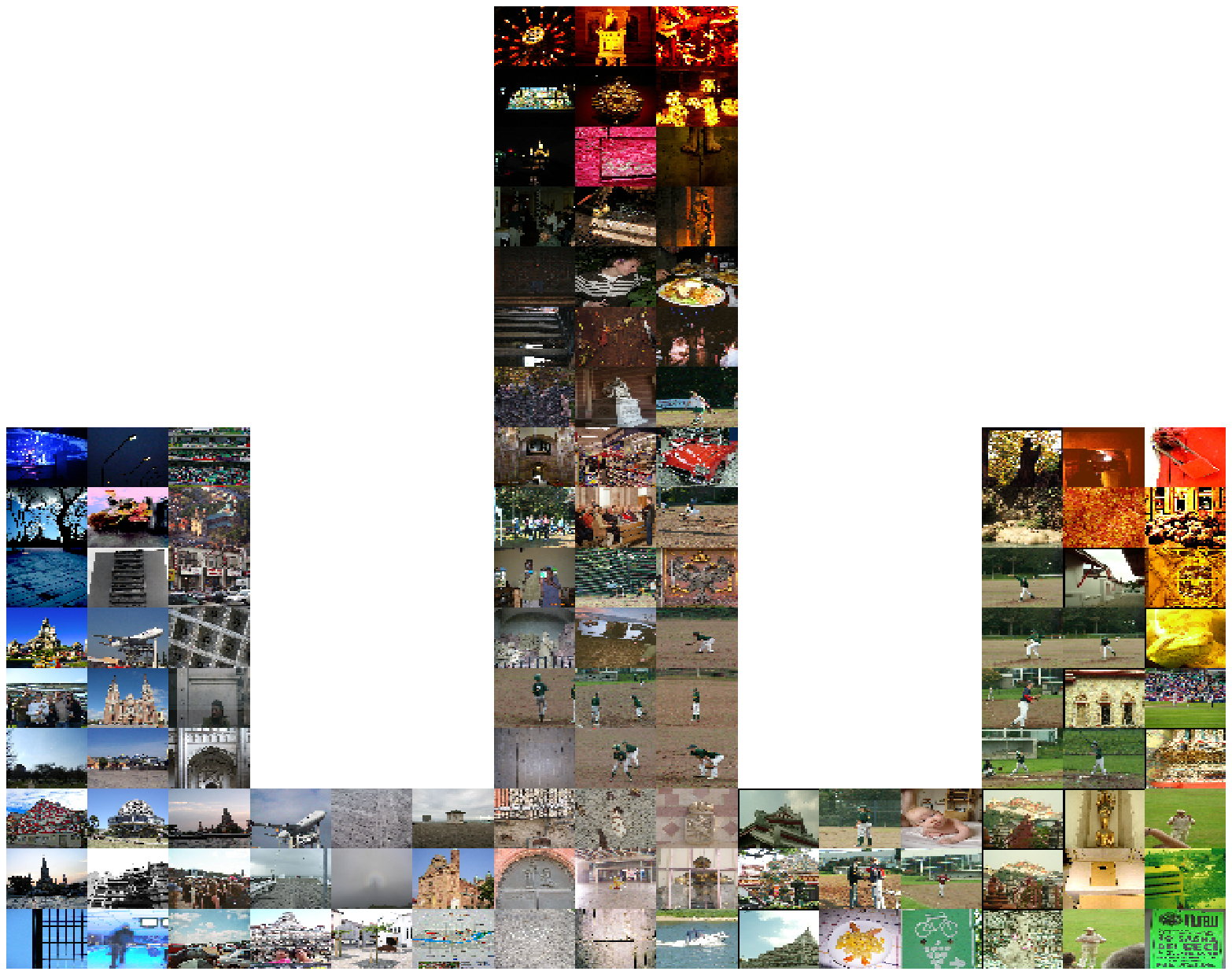}
  \label{fig:image_grid_yama}
}
\hspace*{2mm}
\subfigure[Layout of 153 facial images into `smiley' by LSOM.]{
\includegraphics[width=0.3\textwidth]{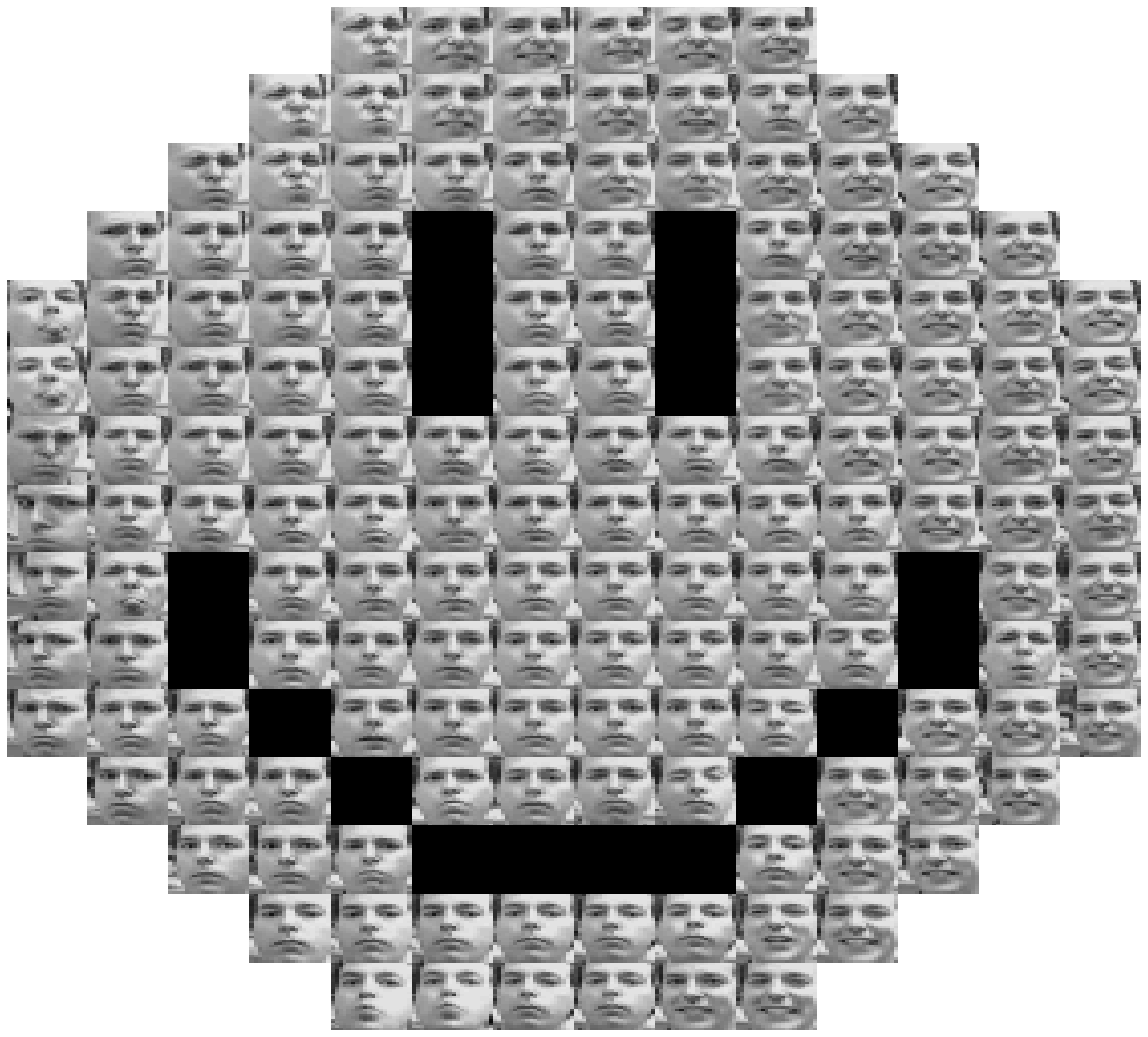}
\label{fig:image_grid_smile}
}
\hspace*{2mm}
\subfigure[Layout of 199 digit `7' into `777' by LSOM.]{
\includegraphics[width=0.3\textwidth]{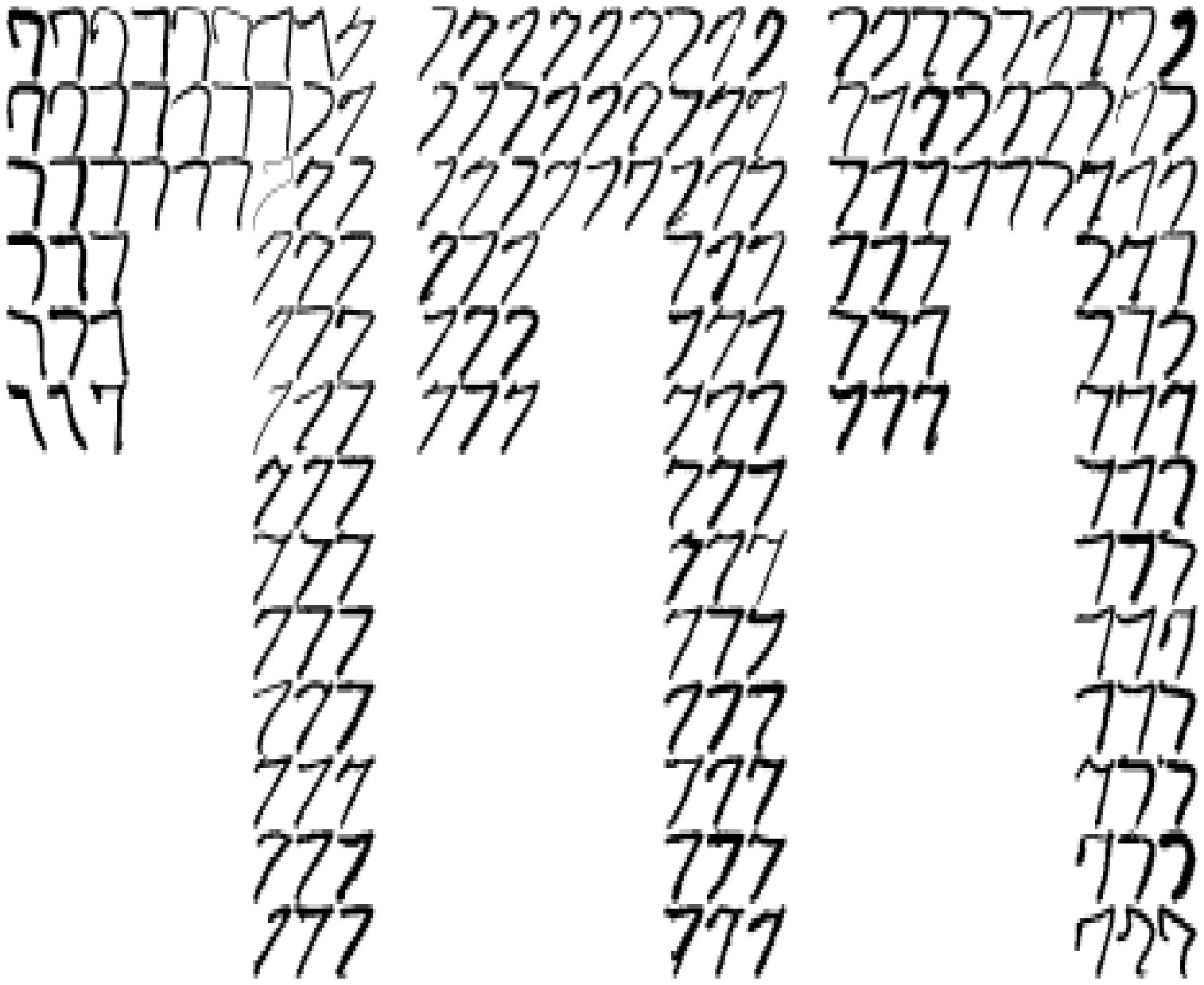}
  \label{fig:usps_seven}
}
 \caption{Images are automatically aligned into complex grid frames expressed in the Cartesian coordinate system.}
    \label{fig:compimages}
\end{figure*}

\subsection{Photo Album Summarization}
Finally, we apply the proposed LSOM method to a photo album summarization problem, 
where photos are automatically aligned into a designed frame expressed
in the Cartesian coordinate system. 

We use 320 images with RGB format used in \citet{PAMI:Quadrianto+etal:2010},
which were originally extracted from \emph{Flickr}\footnote[2]{http://www.flickr.com}.
We first convert the images from RGB to Lab space and resize them to $40 \times 40$ pixels.
Next, we convert a $40 \times 40 \times 3$ $(=4800)$ image into a $4800$-dimensional vector.
We first consider a rectangular frame of $16 \times 20$ $(=320)$,
and arrange the images in this rectangular frame.
Figure~\ref{fig:image_grid} depicts the photo album summarization result,
showing that images are aligned in the way that images with similar colors
are aligned closely.

Similarly, we use the \emph{Frey face dataset}
\citep{Science:Roweis+etal:2000}, which consists of 225 gray-scale face images
with $28 \times 20$ $(=560)$ pixels. We similarly convert a image into a $560$-dimensional vector,
and we set the grid size to $15 \times 15$ $(=225)$. 
The results depicted in Figure~\ref{fig:image_face}
shows that similar face images (in terms of the angle and facial expressions)
are assigned in nearby cells in the grid.

Next, we apply LSOM to the USPS dataset \citep{book:Hastie+Tibshirani+Friedman:2001}.
In this experiment, we use 320 gray-scale images of digit `7' with $16 \times 16$ 
$(=256)$ pixels.
We convert an image into a $256$-dimensional vector, and we set the grid size
to $16 \times 20$ $(=320)$.
The result depicted in Figure~\ref{fig:usps} shows that
digits with similar profiles are aligned closely.

Finally, we align the Flickr, Frey face, and USPS images into more complex frames---a Japanese character `mountain', a smiley-face shape, and a `777' digit shape. The results depicted in
Figure~\ref{fig:compimages} shows that images with similar profiles are located in nearby grid coordinate cells.


\section{Conclusion} \label{sec:conclusion}
In this paper, we proposed two alternative methods 
of cross-domain object matching (CDOM).
The first method uses the dependence measure
based on the normalized cross-covariance operator,
which is advantageous over HSIC in that it is asymptotically independent of the choice of kernels.
However, with finite samples, it still depends on the choice of kernels
which needs to be manually tuned.
To cope with this problem,
we proposed a more practical CDOM approach called \emph{least-squares object matching} (LSOM).
LSOM adopts \emph{squared-loss mutual information} as a dependence measure,
and it is estimated by the method of \emph{least-squares mutual information} (LSMI).
A notable advantage of the LSOM method 
is that it is equipped with a natural cross-validation procedure that allows us to
objectively optimize tuning parameters such as the Gaussian kernel width
and the regularization parameter in a data-dependent fashion. 
We applied the proposed methods to the image matching, unpaired voice conversion,
and the photo album summarization tasks, and experimentally showed
that LSOM is the most promising.

\bibliography{lsom}

\begin{thebibliography}{}
\bibitem[Pearson(1900)Pearson]{PhMag:Pearson:1900}
Pearson, K. (1900).
\newblock On the criterion that a given system of deviations from the probable
  in the case of a correlated system of variables is such that it can be
  reasonably supposed to have arisen from random sampling.
\newblock {\em Philosophical Magazine\/}, {\bf 50}, 157--175.
\end{thebibliography}
\bibliographystyle{natbib}

\end{document}